\documentclass[11pt,a4paper]{article}
\usepackage[hyperref]{emnlp2020}
\usepackage{times}
\usepackage{latexsym}

\usepackage{amsmath}
\usepackage{multirow}
\usepackage{todonotes}
\usepackage{booktabs}
\usepackage{subfigure}
\usepackage{pdflscape}

\usepackage{microtype}

\aclfinalcopy 

\title{Continual Domain-Tuning for Pretrained Language Models }

\author{Subendhu Rongali\thanks{~~Equal contribution}, ~Abhyuday Jagannatha\footnotemark[1], ~Bhanu Pratap Singh Rawat\footnotemark[1], ~Hong Yu \\
  University of Massachusetts Amherst \\
  \texttt{\{srongali, abhyuday, brawat, hongyu\}@cs.umass.edu}}

\date{}

\begin{document}
\maketitle
\begin{abstract}
Pre-trained language models (LM) such as \textsc{bert}, Distil\textsc{bert}, and Ro\textsc{bert}a can be tuned for different domains (domain-tuning) by 
continuing the pre-training phase 
on a new target domain corpus. This simple domain tuning (SDT) technique has been widely used to create domain-tuned models such as Bio\textsc{bert}, Sci\textsc{bert} and Clinical\textsc{bert}. However, during the pretraining phase on the target domain, the LM models may \textit{catastrophically forget} the patterns  
learned from their source domain. In this work, we study the effects of \textit{catastrophic forgetting} on domain-tuned LM models and investigate methods that mitigate its negative effects. We propose continual learning (CL) based alternatives for SDT, that aim to reduce \textit{catastrophic forgetting}. We show that these methods may increase the performance of LM models on downstream \textit{target} domain tasks. Additionally, we also show that constraining the LM model from \textit{forgetting} the \textit{source} domain leads to downstream task models that are more robust to domain shifts. We analyze the computational cost of using our proposed CL methods and provide recommendations for computationally lightweight and effective CL domain-tuning procedures. 

\end{abstract}

\section{Introduction}
Recently proposed pre-trained contextual word embedding and language models (LM) such as \textsc{bert} \citep{devlin2018bert}, \textsc{elm}o \citep{peters2018deep}, \textsc{gpt} \citep{radford2018improving}, \textsc{xln}et \citep{yang2019xlnet} Ro\textsc{bert}a \citep{liu2019roberta}, Distil\textsc{bert} \citep{sanh2019distilbert} and \textsc{albert} \citep{lan2019albert} are widely used in natural language processing tasks. These LM models use unsupervised pre-training to train RNN or Transformer based neural network models on large unsupervised text corpora like WikiText and Gigaword \citep{devlin2018bert,yang2019xlnet}. Simply replacing the hidden layers in an existing neural architecture by a pretrained LM model and fine-tuning the model on a supervised task leads to large performance improvements \footnote{LM model perfromances can be seen at GLUE and SuperGLUE leader-boards. https://super.gluebenchmark.com/, https://gluebenchmark.com/}.

The corpora that LM models are typically trained on consist of text from multiple topics such as news, finance, law, sciences \citep{devlin2018bert,peters2018deep}. Due to their success in NLP applications and their ease of use, researchers have worked on adapting these models for specific downstream domains. We refer to the initial domain as the \textit{source} domain and the downstream domain as the \textit{target} domain. Here, adaptation works as a simple domain transfer technique (SDT) by using the pretrained LM model weights as initialization and continuing the pretraining on a large \textit{target} domain text corpus. The \textit{target} domain LM model is then used as initialization for supervised finetuning on \textit{target} domain downstream tasks. This process has been followed in the biomedical, clinical, and scientific domains to produce Bio\textsc{bert} \citep{lee2019biobert}, Clinical\textsc{bert} \citep{alsentzer2019publicly} and Sci\textsc{bert} \citep{beltagy2019scibert} models. For this paper, we call the pretraining on \textit{source} domain as ``pretraining'', the continued pretraining on \textit{target }domain corpora as ``domain-tuning'', and the final supervised training on the downstream domain-specific task as ``finetuning''.

Domain-tuning through SDT can cause models to \emph{catastrophically forget} \citep{kirkpatrick2017overcoming,mccloskey1989catastrophic} the learned \textit{source} domain information. Continual learning (CL) techniques aim to reduce \textit{catastrophic forgetting} in models that are sequentially trained on an array of tasks. 
Reducing \textit{catastrophic forgetting} may result in \textit{positive forward transfer} (PFT) and \textit{positive backward transfer} (PBT) as demonstrated by CL methods like \citet{kirkpatrick2017overcoming,lopez2017gradient}. Therefore, we study the effects of these CL methods on the domain-tuning procedure for LM models like \textsc{bert}.
In CL, PFT happens when a previous task improves the performance of a later task in the sequence. PBT happens if a later task improves the performance of a previous task in the sequence. 

However, CL concepts like PBT and PFT may not be directly applicable to LM domain-tuning. The domain-tuning task sequence only contains two tasks, the self-supervised pretraining tasks on the \textit{source} and the \textit{target} domain corpora. Additionally, the evaluation of an LM model is not based on its self-supervised loss. Instead, it is based on how ``useful'' that LM model is for a downstream supervised task. So in this context of domain-tuning, we define forward transfer as the effect of \textit{source} domain pretraining on tasks in the \textit{target} domain, and backward transfer as the effect of \textit{target} domain-tuning on tasks in the \textit{source} domain. We hypothesize that retaining more source domain information during domain-tuning may help the LM produce improved representations for both target and source domain texts.  

To exhaustively examine the effect of continual learning methods on LM domain-tuning procedure, we evaluate five different commonly used CL methods. We design an extensive experimental setup to test the PFT and PBT capabilities of these CL domain-tuning methods on three commonly used LM models: \textsc{bert}, Ro\textsc{bert}a, and Distil\textsc{bert}.  We use extractive question answering and natural language inference as our supervised downstream tasks.

Our main findings are as follows:
\begin{itemize}
    \item Different CL methods have varying advantages and disadvantages for domain-tuning depending on model size and transfer direction. We use our experiments to identify and recommend CL methods that adapt well to the task of domain-tuning.
    \item LM models outputs for our recommended CL methods like Elastic Weight Consolidation \cite{kirkpatrick2017overcoming} contain more \textit{information} about downstream task output random variables compared to baseline SDT.
    \item Finetuned downstream task models that are initialized with our recommended CL tuned models are more robust to \textit{target}$\longrightarrow$\textit{source} domain shifts. 
\end{itemize}

Finally, we analyze various CL based domain-tuning methods in the context of their computational cost and discuss their practical benefits. 




\section{Background and Methods}
LM models such as \textsc{bert}, are pretrained on large unsupervised text corpora such as WikiText that contain text from multiple domains. We assume that the text corpora used for \textit{pretraining} is the \textit{source} corpus. \textit{Source} is the first domain to be trained, therefore all CL methods are reduced to standard pretraining. As a result, we can use the published  \textsc{bert}, Distil\textsc{bert} and Ro\textsc{bert}a models as models that are pretrained on our source corpus for SDT and all CL methods. We define the \textit{target} domain and downstream tasks for both \textit{source} and \textit{target} domains in the next section.

Neural network LM models use self-supervised objectives such as masked language modeling (MLM) loss for training. We abstract out the model-specific details and refer to all model losses as $\ell_{domain}(D,\theta)$. Here $\theta$ denotes the parameters of the neural network and $D$ is the unlabeled training dataset. We denote the \textit{source} domain dataset as $D_s$, and the \textit{target} domain dataset as $D_t$. An LM model pretrained on $D_s$ with loss $\ell_{s}$ is domain-tuned on $D_t$ with loss $\ell_{t}$ to obtain a \textit{target} domain LM model. This model can then be used for supervised finetuning on a \textit{target} domain downstream task.

Several of our proposed CL based methods use an additional regularization objective $\mathcal{R}$ along with the domain-tuning loss $\ell_{t}$. The domain-tuned model is obtained by solving the following unconstrained optimization
\begin{equation}
        \operatornamewithlimits{argmin}_{\theta} (\ell_t +
        \lambda \mathcal{R}).
\label{eq:unconstrained}
\end{equation}
Bio\textsc{bert} and other domain-tuned LM models were trained using SDT. Therefore we use SDT as our baseline. SDT's domain-tuning phase only uses the loss $\ell_t$. We describe four proposed CL methods in this section. We also experimented with a distillation based CL method, however due to its high computational cost we only describe and report its results in Appendices \ref{sec:distil} and \ref{sec:add_exp}.

\subsection{Rehearsal (RH)}
\label{sec:rehearsal}
\begin{equation}
        L_{RH} = \ell_t +
        \lambda \ell_s
\end{equation}
Rehearsal scheme \citep{ratcliff1990connectionist} avoids catastrophic forgetting by using a few examples from $D_s$ during domain-tuning on $D_t$. The regularization term in this loss is $\mathcal{R}_{RH}=\lambda \times \ell_s$ to ensure that loss on $D_s$ does not increase significantly. A small subset of $D_s$ data is used for this regularization. This scheme can be interpreted as a multi-task learning model with a smaller weight associated with $D_s$.

\subsection{L2 Regularization (L2)}
\label{sec:L2}
\begin{equation}
        L_{L2} = \ell_t +
        \frac{\lambda}{2} \sum_{i} (\theta_i - \theta^{*}_{s,i})^2
\end{equation}
This scheme \citep{daume2007frustratingly} uses the L2 distance between current parameter values, $\theta$, and the parameter values of the pre-trained \textit{source} domain model, $\theta^{*}_{s}$, as its CL regularization. It imposes an isotropic Gaussian prior on the parameters with $\sigma^2=1$ and $\mu=\theta^{*}_{s}$.
\subsection{Elastic Weight Consolidation (EWC)}
\begin{equation}
        L_{EWC} = \ell_t +
        \frac{\lambda}{2} \sum_{i} F_i(\theta_i - \theta^{*}_{s,i})^2
\label{eq:EWC}
\end{equation}
EWC \citep{kirkpatrick2017overcoming} also imposes a Gaussian prior over the parameters with $\mu=\theta^{*}_{s}$ like L2. However, it uses a diagonal covariance $F$. $F$ here is the diagonal of the Fisher information matrix for the source domain loss $\ell_s$. Therefore, the regularization term in EWC is a weighted L2 distance between parameter value at any given SGD step  $\theta$ and the initial parameter value $\theta^{*}_{s}$. The weight for the $i^{th}$ parameter is given by $F_i$, the $i^{th}$ element in the diagonal. The term $F$ stops parameters that are important for the previous task $D_s$ from changing too much during EWC training. A more detailed explanation of EWC is provided in Appendix \ref{sec:ewc}. 

\subsection{Gradient Episodic Memory (GEM)}

GEM does not use a regularization objective as shown in Equation \ref{eq:unconstrained}. Instead, it poses continual learning as a constrained optimization problem. In our context, the problem aims to minimize $\ell_t$ subject to $\ell_s(D_s,\theta) \leq \ell_s(D_s;\theta^*)$. Recall that $\ell_s$ and $\ell_t$ refer to losses on the \textit{source} and \textit{target} domain corpora. GEM solves this constraint optimization problem by making local corrections to gradients at each SGD step. 

GEM assumes that the loss function is locally linear at each SGD step. As a result, we can rephrase the previous constrained optimization problem as that of finding a gradient $g$, which is very close to the target domain gradient $\frac{\partial \ell_t}{\partial \theta}$ and does not have any negative component in the direction of source domain gradient $\frac{\partial \ell_s}{\partial \theta}$. Since we only have one constraint in our formulation, we can analytically solve for $g$ instead of using a quadratic program solver as suggested by GEM. Details about the solution are provided in Appendix \ref{sec:gem}.



\section{Experimental Design}
We collect \textit{source} domain and \textit{target} domain corpora based on related efforts in literature. We use the WikiText corpus, which is a large common part of the datasets used for pretraining \textsc{bert}, Ro\textsc{bert}a and Distil\textsc{bert}, as our \textit{source} domain unlabeled corpus $D_s$. We perform our experiments over three different variants of transformer \citep{vaswani2017attention} based language models: \textsc{bert} \citep{devlin2018bert}, Ro\textsc{bert}a \citep{liu2019roberta} and Distil\textsc{bert} \citep{sanh2019distilbert}.

\subsection{\textit{Target} domain unlabeled corpora}
Our selection of target domains was dictated by the availability of the unlabeled text corpus and standard downstream tasks for that domain. For our primary analysis, we construct a \textit{target} domain text corpora containing various biomedical and clinical text documents. We also evaluate our methods on a smaller clinical only \textit{target} domain corpus and the results on this second corpus along with further details about corpus selection are included in Appendix \ref{sec:appendix_b} and \ref{sec:add_exp}. 

The bio-medical text is obtained by crawling through a snapshot of all the abstracts of papers made available by PubMed Central (PMC)\footnote{\url{https://www.ncbi.nlm.nih.gov/pmc/}}. 
For the clinical text, we extract all the clinical notes from the MIMIC-III dataset \citep{mimiciii}, which consists of electronic health records of patients who stayed within the intensive care units at Beth Israel Deaconess Medical Center. This corpus is identical to the one used by \citet{alsentzer2019publicly} and a superset of \citet{lee2019biobert}. 

\subsection{Domain Tuning Details}
For evaluating all our approaches, we perform domain-tuning by allowing one full iteration over the \textit{target} domain corpus \footnote{Code and trained model parameters available at https://anonymized}. Further experimental details are provided in Appendix \ref{sec:appendix_a}.

\subsection{Downstream Tasks} We use two widely used NLP tasks, Question-Answering (QA) and Natural Language Inference (NLI) as our downstream tasks. These tasks have domain specific datasets for both \textit{source} and \textit{target} domains. For the source domain, we pick SQuAD 2.0 \citep{rajpurkar2018know} and SNLI \citep{bowman2015large} datasets. For the \textit{target} domain, we use EMRQA \citep{pampari2018emrqa} \footnote{The emrQA dataset was transformed into SQuAD-style and examples which could not be transformed were removed.} and MedNLI \citep{romanov2018lessons}. We run each downstream experiment three times and report the mean values in our tables. Tables with standard deviations and relevant implementation details are provided in the Appendices \ref{sec:appendix_a}, \ref{sec:appendix_b} and \ref{sec:add_exp}.

\begin{table*}
\centering
\resizebox{\linewidth}{!}{%
\begin{tabular}{cc|ccc|ccc}
\hline
& & \multicolumn{3}{c|}{\textbf{Downstream tasks related to \textit{Source} domain}} & \multicolumn{3}{c}{\textbf{Downstream tasks related to \textit{Target} domain}}\\
\hline
& & \multicolumn{2}{c}{\bf SQuAD 2.0} & \bf SNLI & \multicolumn{2}{c}{\bf EMRQA} & \bf MedNLI\\
\hline
& & EM & F-score & Accuracy & EM & F-score & Accuracy\\
\hline
\multirow{6}{*}{\rotatebox{90}{\textsc{bert}}}
& \bf NDT & 71.78 & 75.50 & 90.45& 74.86 & 80.43 & 78.49\\
& \bf SDT & 71.25 & 74.63 & 88.45 & \textbf{76.3}	& \textbf{81.86} & 79.92\\
& \bf RH & 70.76(-0.49)&74.41(-0.22)&89.17(0.72)&\textbf{76.26(	-0.04)}&\textbf{81.86(0)}	&80.46(0.54)\\
& \bf EWC & 71.11(-0.14)&74.81(0.18)&90.14(1.69)&75.8(-0.5)&81.29(-0.57)&\textbf{80.93(1.01)}\\
& \bf GEM &  68.88(-2.37)&72.43	(-2.2)&88.37	(-0.08)&75.9(-0.4)&	81.71	(-0.15)&	78.47	(-1.45)\\
& \bf L2 &  \textbf{72.18(0.93)} &\textbf{75.86(1.23)}&\textbf{90.4(1.95)}&75.76(-0.54)&80.96(-0.9)&78.14	(-1.78)\\
\hline
\multirow{6}{*}{\rotatebox{90}{Ro\textsc{bert}a}} 
& \bf NDT & 78.16&	81.79&	91.38&	74.57&	80.45&	81.04	\\
& \bf SDT & 72.61&	76.34&	89.59&	74.24&	80.09&	83.98	\\
& \bf RH & 75.61(3.00)&79.31(2.97)&90.75(1.16)&74.93(0.69)&80.7(0.61)&\textbf{85.43	(1.45)}\\
& \bf EWC & 77.67(5.06)&81.44(5.1)&91.3(1.71)&\textbf{74.98	(0.74)}&\textbf{80.81(0.72)}&83.65	(-0.33)\\
& \bf GEM & 72.75(0.14)&76.57(0.23)&89.87(0.28)&74.27	(0.03)&80.11(0.02)&84.43	(0.45)\\
& \bf L2 &  \textbf{78.25(5.64)}&\textbf{81.84(5.5)}&\textbf{91.44(1.85)}&74.68	(0.44)&80.47	(0.38)&80.44	(-3.54)\\
\hline
\multirow{6}{*}{\rotatebox{90}{Distil\textsc{bert}}}
& \bf NDT &65.47	&69.18&	89.69&	73.56&	79.3&	76.07	\\
& \bf SDT & 64.88	&68.4&	87.25&	74.21&	\textbf{80.26}&	78.27	\\
& \bf RH & 65.14	(0.26)&68.89	(0.49)&88.87	(1.62)&\textbf{74.35	(0.14)}&80.2	(-0.06)&78.46	(0.19)\\
& \bf EWC & 65.22(0.34)&\textbf{69.01(0.61)}&89.47(2.22)&73.71(-0.5)&79.5(-0.76)&76.26(-2.01)\\
& \bf GEM &  64.17(-0.71)&67.78(-0.62)&87.98(0.73)&74.26(0.05)&\textbf{80.05(-0.21)}&\textbf{78.58(0.31)}\\
& \bf L2 &  \textbf{65.27(0.39)}&68.92(0.52)&\textbf{89.67(2.42)}&73.57(-0.64)&79.38(-0.88)&75.81(-2.46)\\
\hline
\end{tabular}
}
\caption{\label{tab:results}
Downstream task results from our experiments across all models and techniques for \textit{target} domain. LM models are pretrained on \textit{source} domain, CL domain-tuned on \textit{target} domain and finetuned on each downstream task to obtain the results. All results are averaged over three runs. NDT refers to the base model that is not domain tuned.}
\end{table*}
\subsection{Experiments for Forward Transfer}
We investigate our hypothesis for \textit{positive forward transfer}, that retaining \textit{source} domain information during domain-tuning may result in ``better adapted'' LM models for downstream \textit{target} domain tasks, in two ways: 1) We use the \textit{target} domain-tuned models as initialization for finetuning on \textit{target} domain downstream tasks. An improved test performance of the trained downstream task can indicate better adaptation. 2) We examine the hidden layer and output layer representations of the CL domain-tuned models. This may also provide indications of better adaptation. Specifically, we examine the mutual information between the model's output and output random variable of the downstream task. 

\subsection{Experiments for Backward Transfer}
We design two experiments to evaluate \textit{positive backward transfer}. The first experiment uses the \textit{target} domain-tuned model as initialization for training downstream \textit{source} domain tasks. Additionally, we also hypothesize that retaining \textit{source} domain information may result in downstream target domain models that are more robust to \textit{target}$\longrightarrow$\textit{source} domain shifts. To test this hypothesis, we study the performance of EMRQA and MedNLI finetuned models on \textit{source} domain SQuAD and SNLI tasks, respectively.

\begin{figure*}[h]
\centering
\includegraphics[width=1.0\textwidth]{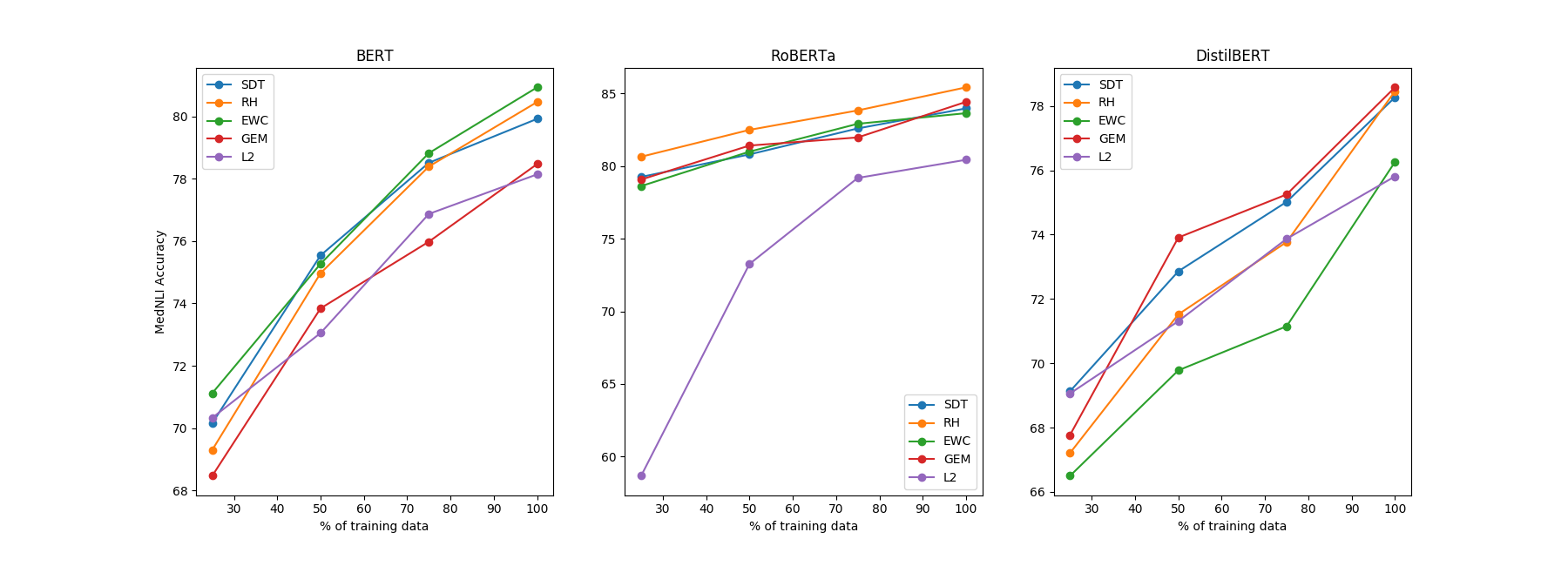}
\vspace{-2em}
\caption{MedNLI test set accuracy (averaged over three runs) of models with varying training set size.}
\label{fig:datase_size}
\end{figure*}

\begin{table}
\begin{tabular}{cccccc}
\hline
 Model & EWC & GEM & L2 & RH \\ \hline \hline
 BERT & \textbf{+1.71} & +0.71 & -1.71 &  -0.25 \\
 DistilBERT & \textbf{+1.26} & -0.50&  -1.38& +0.25 \\
 RoBERTa & +1.46 & \textbf{+2.52} & +1.14& +1.33 \\ \hline
 Avg & \textbf{+1.48} & +0.91 &  -0.65     &   +0.44 \\ \hline
\end{tabular}
\caption{\label{tab:mi}
This table shows  $(\mathcal{I}(y;\phi_{CL})-\mathcal{I}(y;\phi_{SDT}))\times 10^2$ for CL models. Values are averaged over three runs.}
\end{table}

\begin{table*}
\centering
\begin{tabular}{cccccc}
\hline
 Model & EWC & GEM & L2 & SDT & RH \\ \hline \hline
 B (BERT) & 31.77/21.07 & 29.74/19.46 & 30.54/19.50 & 30.71/20.34 & \textbf{32.50/22.23} \\
 R (RoBERTa) & 33.08/22.21 &27.50/17.07&  \textbf{34.97/23.80} &25.00/15.17& 30.07/19.49 \\
 D (DistilBERT) & 25.53/16.34 & 23.43/13.85 &24.31/15.06& 22.20/13.20 & \textbf{26.35/16.56} \\ \hline
  Avg & \textbf{30.13}/\textbf{19.87} &26.89/16.79 &29.94/19.45 & 25.97/16.24& 29.64/19.42  \\ \hline
\end{tabular}
\caption{F1-score and Exact Match (F1/EM) of all models (averaged over three runs) trained on EMRQA training set and evaluated on the answerable only validation set of SQuAD 2.0. All models trained using EWC and RH perform better than SDT.}
\label{tab:emrqa_squad}
\end{table*}

\begin{table}
\begin{tabular}{cccccc}
\hline
 LM & EWC & GEM & L2 & SDT & RH \\ \hline \hline
 B & \textbf{41.13} & 38.01 & 39.89 & 39.56 & 39.16 \\
 R & 51.18 &42.29&  43.28&45.05& \textbf{51.75} \\
 D & 39.80 & 37.81 &\textbf{41.19}& 37.83 & 39.43 \\ \hline
  Avg & \textbf{44.03} & 39.37 & 41.45 & 40.81& 43.45\\ \hline
\end{tabular}
\caption{Accuracy (averaged over three runs) of models trained on MedNLI and evaluated on SNLI test set.}
\label{tab:mnli_snli}
\end{table}

\section{Experiments and Results}
\subsection{Forward Transfer}
\paragraph{Downstream Performance}
We finetuned all domain-tuned LM models on EMRQA and MedNLI tasks for the \textit{target} domain. The results for \textit{target} domain downstream performances are present in right half of Table \ref{tab:results}. We see that in several cases, the best performing CL model outperforms the baseline SDT model. RH appears to be the best performing continual learning model in terms of downstream performance for \textit{target} domain tasks. GEM and EWC appear to be the next best performing CL models. 

The high performance of RH model can be explained by its use of both $D_s$ and $D_t$ losses in its objective. RH is effectively a multi-task learning approach, that is simultaneously trained on both source and target domain samples during domain-tuning. Unlike EWC or L2, it does not introduce explicit model parameter constraints, and therefore it does not suffer from the problem of over-constraining the model. As a result, it has better source domain regularization. EWC and GEM show mostly competitive or improved performance as compared to SDT, except for a few instances. EWC tuned models show significant regressions in Distil\textsc{bert}. We believe this is due to the smaller size of Distil\textsc{bert} compared to other models. EWC is known to over-constrain the parameters in a continual learning setup, preventing the network from learning new information. This aspect may more severely affect networks with smaller number of parameters. Therefore EWC performance is lower in Distil\textsc{bert} models. L2, which over-constrains the parameters even more than EWC, leads to a larger decrease in downstream performance. 

\paragraph{Downstream Performance Analysis on Varying Data-set Lengths}
We see a more pronounced trend of model difference when we study the downstream results with varying labeled training data sizes in Figure \ref{fig:datase_size}. We use MedNLI for the varying dataset size experiment. Each domain-tuned model is finetuned on 25\%, 50\%, 75\% and 100\% of the MedNLI task training set. In the case of \textsc{bert} models, EWC and RH remain competitive to or are better than SDT's performance. We see similar behavior for Ro\textsc{bert}a. Due to the aforementioned over-constraining issue in EWC and L2, we see reduced performances from them for Distil\textsc{bert}. GEM and RH are still competitive for Distil\textsc{bert}.

\paragraph{Analysis of layer outputs in LM models}
To further analyze CL domain-tuned models for the target domain, we analyse the quality of each model's hidden representation in the context of downstream \textit{target} domain task. We use MedNLI dataset for these sets of experiments. To estimate the quality of hidden representation, we use mutual information (MI) $\mathcal{I}(Y;\phi_{(.)}(X))$ between a model's hidden representation $\phi_{(.)}(X)$ and the output variable of MedNLI, $Y$ . Table \ref{tab:mi} shows the difference in mutual information $\mathcal{I}(Y;\phi_{(.)}(X))-\mathcal{I}(Y;\phi_{SDT}(X))$ for our proposed CL methods. Here $\phi_{(.)}(X)$ denotes the concatenation of all hidden layer and final layer output representations of a domain-tuned (but not fine-tuned) model. Details of MI estimation are provided in Appendix \ref{sec:app_mi}.

From Table \ref{tab:mi}, we see that hidden representations obtained from CL trained models like EWC, GEM, RH have higher mutual information with downstream task outputs. This observation further supports our hypothesis that CL domain-tuning may result in \textit{positive forward transfer}. EWC model has the highest mutual information amongst all CL models, followed by RH and GEM.

\subsection{Backward Transfer }
\paragraph{Downstream Task Performance}
The left half section of Table \ref{tab:results} shows \textit{source} task downstream results for \textit{target} domain-tuned models that were finetuned on \textit{source} domain downstream tasks. We see that L2 outperforms all other models on \textit{source} domain tasks. This is mainly due to the heavy constraints that L2 puts on model parameters. These constraints reduce the model adaptation on \textit{target} domain. As a result, L2 performs worst on \textit{target} domain and best on \textit{source} domain. EWC, RH and GEM are the next best models in order.

\paragraph{Performance Under Domain Shift}
Ensuring that LM models retain their knowledge about the \textit{source} domain may also contribute to making our downstream task models more robust. For instance, a CL trained LM model that is finetuned for MedNLI task may produce an NLI model that is more robust to domain shifts. This is especially true if the domain shift is towards the \textit{source} domain.  To test this hypothesis, we used models finetuned on target domain downstream tasks (Right half of Table \ref{tab:results}) and evaluated them on the same task's source domain variant. So models trained on EMRQA were evaluated on SQuAD, and those trained on MedNLI were evaluated on SNLI. Tables \ref{tab:emrqa_squad} and \ref{tab:mnli_snli} show the results for these experiments. RH and EWC models show the best performance in this evaluation. These models show large improvements against the baseline SDT in both EMRQA and MedNLI tasks. Surprisingly, the L2 method that has the strongest constraints against model adaptation does not have the best performance in these experiments. GEM also has reduced performance compared to EWC and RH.

\begin{figure}[ht]
\centering
\includegraphics[width=0.40\textwidth]{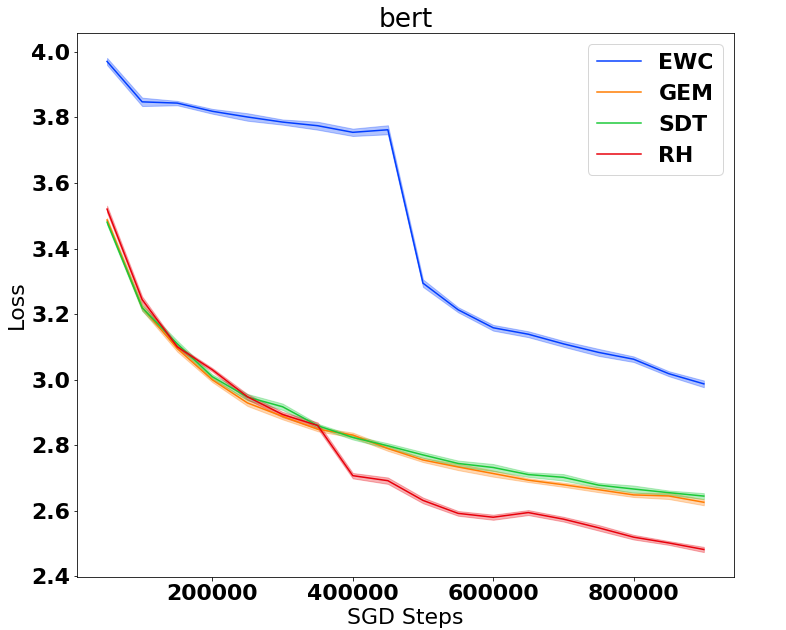}
\caption{Self-supervised loss $\ell_t$ on \textit{target} domain validation set for \textsc{bert} CL models. L2 model's loss for \textsc{bert} is very high, and is not included in the graph to show higher y-axis resolution for other CL models.}
\label{fig:bio_perplexity_bert}
\end{figure}

\begin{figure}[h]
\centering
\includegraphics[width=0.40\textwidth]{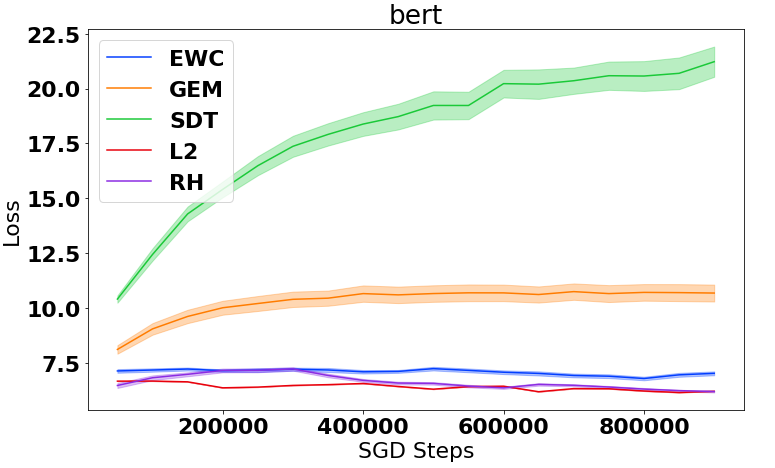}
\caption{Self-supervised loss $\ell_s$ on \textit{source} domain validation set for \textsc{bert} CL models.}
\label{fig:wiki_perplexity_bert}
\end{figure}

\subsection{Self-Supervised Loss Analysis}
Figures \ref{fig:bio_perplexity_bert} and \ref{fig:wiki_perplexity_bert} show self-supervised loss (MLM loss) values during domain-tuning for \textit{target} and \textit{source} domains for \textsc{bert}\footnote{Figures for Ro\textsc{bert}a and Distil\textsc{bert} are provided in Appendix.}. Loss values were averaged over 5 subsets of 50\% randomly selected samples from validation dataset. \textit{Source} domain (Figure \ref{fig:wiki_perplexity_bert}) loss values show that CL domain-tuning methods preserve low loss values on the source domain throughout the domain tuning process.

From Figure \ref{fig:bio_perplexity_bert}, we see that GEM and RH are close to, or have lower loss values that SDT on the \textit{target} domain. EWC and L2 have higher loss values compared to SDT, which seems to contradict the \textit{forward transfer} results. An explanation for this may be the over-constrained last layer. Fast adaptation of the final softmax layer is crucial in reducing the \textit{target} MLM loss. Some continual learning works use separate tasks-specific last layer \citep{li2017learning}, but we keep it same to ensure comparison with SDT. Both EWC and L2 impose constraints over model parameters which may restrict the adaptation of last layer. EWC, in particular, may suffer from slow adaptation of the last layer,  because last layer may have high empirical Fisher estimates. As a result, the large precision matrix entries for last layer may result in very slow adaptation of last layer, even when hidden layers of the model are rapidly adapting to the new \textit{target} domain. This may result in slow \textit{target} domain self-supervised loss decrease even though the hidden layers of the network are adapting well to the final domain. We examine the Fisher information estimates and find that last layer parameters do have higher constraints. For instance, \textsc{bert} final layer Fisher estimates are two orders of magnitude higher than those of the 12 transformer layer parameters. In GEM and RH, CL constraints are not explicitly enforced on the model parameters or the last layer. Therefore, they do not suffer from this phenomenon. RH has the best decrease in \textit{target} loss, which aligns with our \textit{forward} transfer results.

\section{Discussions and Recommendations}
In the previous sections, we use multiple experimental designs to isolate the effect of CL based domain-tuning methods in the context of \textit{backward and forward transfer} for LMs.

\paragraph{Backward Transfer}
In our domain shift experiments, we see strong evidence suggesting that backward transfer may make downstream task models more robust to \textit{target}$\longrightarrow$\textit{source} domain shift. They show one of the main practical benefits of CL based domain tuning. Most real-world applications have to deal with heterogeneous text consisting of sentence samples from multiple topics or sub-domains. For instance, narrative notes from physicians, or medical question-answering chat-bot responses may contain sentences that are outside their specific sub-domains. Since the \textit{source} domain in most pretrained models is a collection of texts from a diverse range of topics (such as WikiText or Common Crawl Corpus), our exhibited domain robustness may be useful in a variety of such real-world applications.

\paragraph{Forward Transfer and Model Capacity}
Our \textit{forward transfer} experiments show that models with larger size like \textsc{bert} and Ro\textsc{bert}a have the capacity to effectively domain-tune on \textit{target} domain text even when they are constrained by CL objectives. Both \textsc{bert} and Ro\textsc{bert}a show performance similar to the SDT model on our downstream tasks. For CL methods EWC, RH and GEM both \textsc{bert} and Ro\textsc{bert}a models either show improvement in performance or are close to SDT performance, while retaining \textit{source} domain information. For smaller models such as Distil\textsc{bert}, CL methods like L2 and EWC prove to be overly tight constraints. We see that RH and GEM, lead to improved downstream task performance of Distil\textsc{bert} models. For both GEM and RH, Distil\textsc{bert} models are competitive to baseline SDT for EMRQA and have higher than baseline performance on MedNLI. This is because RH and GEM do not enforce an explicit constraint on model parameter values and therefore do not over-constraint the lower capacity Distil\textsc{bert} model.

\paragraph{Computational Complexity}
In terms of computational complexity, our best model is L2. It only requires an additional squared norm objective in the loss function and is therefore only marginally more computationally expensive for auto-differentiation libraries like PyTorch. EWC has similar computational complexity as L2, with the added one-time overhead of estimating diagonal entries of the empirical Fisher matrix. We estimate these entries using a small section of the source WikiText dataset. 
Since the Fisher matrix is only computed once during the entire domain-tuning run, EWC only increases the computational cost by a small fraction. Both L2 and EWC methods are almost similar to SDT in terms of computational complexity. 

RH requires two mini-batch computations for every SGD iteration and is therefore almost twice as computationally expensive as SDT. GEM, which computes gradients for the \textit{source} domain at every SGD mini-batch, also has higher memory requirements. In our experiments, we use the same mini-batch size for \textit{source} domain and \textit{target} domain. As a result, in our setup, RH and GEM are twice as expensive as SDT, EWC and L2. Reducing the batch size for \textit{source} domain will reduce the computational cost at the expense of more noisy mini-batch gradient estimates. Distillation \citep{hinton2015distilling} CL runs require three times the computational cost as SDT, and therefore we do not use it in our main results. Further explanation is provided in Appendices \ref{sec:distil} and \ref{sec:add_exp}.

\paragraph{Recommendations}
We have shown that RH and EWC  domain-tuning methods result in the best backward and forward transfer performance for larger capacity models like \textsc{bert} and Ro\textsc{bert}a. RH has better forward transfer results on average, and EWC edges ahead in the backward transfer results. One major deciding factor between the two is the significantly lower computational complexity of EWC. As a result, we recommend using EWC for domain-tuning. It leads to better latent representation and more robust downstream models as compared to SDT. It also leads to competitive or better forward transfer performance as compared to SDT. EWC does not show good performance on a smaller capacity model such as Distil\textsc{bert} due to reasons discussed previously. GEM and RH are likely candidates for CL based domain-tuning for Distil\textsc{bert}. They both have better forward transfer than EWC and are better or competitive as compared to SDT results. However, these methods are more computationally expensive than EWC. 

\paragraph{Open Questions and future work.}
In this work, we investigated continual learning in the context of domain-tuning. Continual learning can be easily extended to include downstream target tasks. CL based downstream task finetuning has the potential to improve both performance and robustness on downstream tasks. The CL methods themselves can also be modified to better accommodate the challenges of LM domain-tuning. For instance, GEM \textit{source} domain gradient constraints are calculated for each mini-batch. These estimates can have very high variance in domain-tuning when the domain corpus is very large. Efficient variance reduction schemes for gradients may improve GEM domain-tuning performance. 

\section{Related Work}
\paragraph{Domain-Specific Pretraining:} The task of adapting pretrained language models for a specific domain is popular in literature. Bio\textsc{bert} \citep{lee2019biobert} and Clinical\textsc{bert} \citep{alsentzer2019publicly} were built for the bio-medical domain from a baseline \textsc{bert} model by finetuning on a biomedical corpus consisting of PubMed articles and the MIMIC-III dataset, just like ours. However, these models are finetuned by simply training further on new text. Sci\textsc{bert} \citep{beltagy2019scibert} is also trained in this fashion but also has its own domain-specific vocabulary. We train simple domain transfer models on our data as one of our baselines.
\paragraph{Continual Learning:} Several works in continual learning focus on overcoming \emph{catastrophic forgetting} when learning on new tasks. Methods like L2 \citep{daume2007frustratingly}, EWC \citep{kirkpatrick2017overcoming}, Variational Continual Learning \citep{nguyen2017variational}, and Synaptic Intelligence \citep{zenke2017continual} use different regularization approaches to constrain the training on new tasks. Methods like Progress and Compress \citep{schwarz2018progress} use modifications to the neural network architecture to increase the capacity of the neural network for a new task. 
We also explore GEM \citep{lopez2017gradient}, a CL approach that was previously shown to outperform EWC on synthetic vision-based continual learning tasks.

\section{Conclusion}
We have investigated the effects of \textit{catastrophic forgetting} on domain-tuning of large pretrained LMs. We proposed CL based domain-tuning methods that constrain the LM from \textit{catastrophically forgetting} the \textit{source} domain while maintaining competitive or improved performance on \textit{target} downstream tasks. We show that our methods create downstream task models that are more robust to domain shifts. In conclusion, we identify and recommend domain-tuning methods like EWC, that have beneficial \textit{forward and backward transfer} properties, while only marginally increasing the computational cost.



\bibliography{emnlp2020}
\bibliographystyle{acl_natbib}

\clearpage
\appendix
\section{Domain-Tuning}
\subsection{Distillation}
\label{sec:distil}
Distillation is a technique used to distil the knowledge from one neural network to another \cite{hinton2015distilling}. It was used in Distil\textsc{bert} \cite{sanh2019distilbert} to train a smaller model by distilling the knowledge from a full \textsc{bert} model into the smaller model. We use distillation as a domain adaptation technique by distilling the knowledge of the pre-trained \textit{source} domain model, denoted by $M_s$, into the domain-tuned model, $M_t$, while domain-tuning with the \textit{target} domain corpus. To do this, we use the distillation loss as our regularization term $\mathcal{R}$. 
\begin{equation}
        L_{DIS} = \ell_t +
        \frac{\lambda}{2} L_{distillation}
\label{eq:dis}
\end{equation}

In neural networks with output classification layers, the distillation loss corresponds to a cross entropy loss between the final output logits of the model being trained and the source model who's knowledge is being distilled into the model. In the scenario of domain tuning, the output logits correspond to the masked token prediction in the MLM objective. So we want our domain-tuned model, $M_t$'s logit distribution on a sample of \textit{source} domain text, $d_s$, to match that of the pre-trained \textit{source} domain model $M_s$. The distillation loss is hence given by
\begin{equation}
    L_{distillation} = CE(M_s(d_s), M_t(d_s))
\end{equation}

This technique is computationally very expensive. For each loss computation, we have the traditional forward pass of the current model on a \textit{target} domain batch, and two additional forward passes, one of the current model on a \textit{source} domain batch, and one of the \textit{source} domain pre-trained model on the \textit{source} domain batch. Because of this huge cost, we leave this method out of our main paper.

\section{Implementation Details}
\label{sec:appendix_a}

\subsection{Domain-tuning}

We used standard values from the huggingface library \citep{Wolf2019HuggingFacesTS} for hyper-parameters such as learning rate, warm-up steps etc. for each of our three BERT variants. Due to the breadth of our experiments and associated computational costs, we were unable to perform hyper-parameter tuning. The CL approaches all have varying memory footprints but we fixed the batch-size to 8 (7 for clinical-only domain-tuning dataset), with each instance in the batch containing 512 tokens, to ensure a fair comparison across them. 

There were also a fair amount of CL method-specific hyper-parameters and we chose the recommended values from the original implementations of these methods wherever possible. Here, we note the values we changed. The loss multiplier in EWC was set to 1e-4 and that of L2 was set to 1. The GEM updates were performed in an online fashion as described in \citet{lopez2017gradient}. For rehearsal and distillation, we need access to data from the \textit{source} domain, so we sample another mini-batch of 512-token instances from the WikiText corpus. We use a multiplier of 0.1 on the source batch loss in these two approaches.

To train models using Masked LM (MLM) objective, we create instances containing contiguous spans for text from the text corpus and mask a small random percentage of words in them. Following \textsc{bert}'s implementation detail, we also mask 15\% of the words. The model is then trained to predict these masked out words using all the visible words. For \textsc{bert} we also used a next sentence prediction (NSP) objective which enables the model to learn language inference features by tasking the model to differentiate between two continuous spans of text and two randomly chosen spans of text. Ro\textsc{bert}a has shown that the NSP objective can be removed without affecting the performance of the overall model and hence doesn't use NSP prediction objective for training. We use a default learning rate of $5e-5$ for all domain-tuning and fine-tuning experiments. Our batch-size for domain-tuning was either 7 or 8 depending on GPU availability.

Our experiments are based on Huggingface's PyTorch library \citep{Wolf2019HuggingFacesTS}. We use the publicly released model weights as our pretrained models\footnote{  bert-base-uncased, roberta-base, and distilbert-base-uncased from \url{https://github.com/huggingface/transformers}}. We use titanx and 1080ti GPUs for our domain-tuning. We ran domain-tuning for one epoch on the \textit{target} dataset. Total time for one epoch ranged from one week to four weeks depending on the \textit{target} dataset size and the computation requirements of the CL method.

The $\lambda$ values used in domain-tuning are 
$1.0$, $1e4$, $0.1$ and $0.1$ for L2, EWC, DIS and RH respectively. GEM does not require a $\lambda$ hyper-parameters. We do not run a hyper-parameter grid search due to computational cost constraints. 

\subsection{Downstream Finetuning}
For all SNLI and MedNLI downstream tasks, we used the default hyper-parameters provided by \url{https://github.com/huggingface/transformers} for NLI tasks. For SQuAD and emrQA, all the default parameters are kept same except the maximum number of training epochs. For emrQA, all models are run for $5$ epochs. For SQuAD,  \textsc{bert} and Ro\textsc{bert}a are run for 2 epochs whereas Distil\textsc{bert} is only run for $1$ epoch.

SQuAD and emrQA were finetuned with a total batch-size of 24. MedNLI and SNLI were trained using a larger total batch-size of 40, since they have shorter text sequences. All these methods were trained on 2 GPUs (1080ti, TitanX or M40).

\section{Methodological Details}
\label{sec:app_method}

\subsection{Elastic Weight Consolidation (EWC)}
\label{sec:ewc}
The regularization term in EWC is a weighted L2 distance between parameter value at any given SGD step  $\theta$ and the initial parameter value $\theta^{*}_{s}$. The weight for the $i^{th}$ parameter is given by $F_i$. $F_i$ in Equation \ref{eq:EWC} is the $i^{th}$ element in the diagonal Fisher estimate.

The EWC regularization penalty is inspired by bayesian online learning. EWC uses a laplace approximation of posterior distribution of $\theta$ obtained from previous tasks as a prior for the current task. In our framework it translates to $\log P(\theta | D_{s},D_{t}) \propto \log P(D_t|\theta) + \log P(\theta|D_{s})$. The procedure for elastic weight consolidation \citep{kirkpatrick2017overcoming} approximates $\log P(\theta|D_{s})$ by using Laplace approximation \citep{mackay1992practical}. Intuitively, the term $\log P(\theta|D_{s})$ denotes information about the weights $\theta$ in the context of the \textit{source} dataset $D_s$. \citet{kirkpatrick2017overcoming} remark that this information refers to which parameter values are important for the previous task. 

The term $\log P(\theta|D_{s})$ is approximated by a Guassian with mean $\theta^{*}_s$ and diagonal precision $F$. The parameters $\theta^{*}_s$ are produced by estimating the Empirical Fisher Information on $D_s$. The EWC term stops parameters that are important for the previous task $D_s$ from changing too much during EWC training. For more details, refer to \citet{kirkpatrick2017overcoming}.

\subsection{Gradient Episodic Memory}
\label{sec:gem}
Here, we describe our solution to obtain the GEM gradient for our CL experiments. Since we only have one constraint in our formulation, we can analytically solve for $g$ instead of using a quadratic program solver as suggested by GEM. We first check whether the \textit{target} domain gradient has a negative component in the direction of \textit{source} domain gradient. This is done by checking if the value of the inner product $<\frac{\partial \ell_t}{\partial \theta}, \frac{\partial \ell_s}{\partial \theta}>$ is negative. If the inner product is positive, we can step in the direction of $\frac{\partial \ell_t}{\partial \theta}$ (within the local linear neighborhood) without increasing the \textit{source} domain loss. If the inner product is negative, we subtract the negative component of $\frac{\partial \ell_t}{\partial \theta}$ along $\frac{\partial \ell_s}{\partial \theta}$ to obtain a gradient $g$. The resulting gradient vector $g$ has a non-negative inner product with $\frac{\partial \ell_s}{\partial \theta}$ and is used for the SGD model update.

\subsection{Mutual Information}
\label{sec:app_mi}
Mutual information $\mathcal{I}(Y;\phi_{(.)}(X))$ can be decomposed as 
\begin{equation}
\begin{split}
    \mathcal{I}(Y;\phi_{(.)}(X)) =\mathcal{H}(Y) - \\ 
    E_{P(\phi_{(.)}(X),Y)}[\log P(Y|\phi_{(.)}(X))]
\end{split}
\end{equation} 
The second term is obtained by rewriting the conditional entropy $\mathcal{H}(Y|\phi_{(.)}(X))$ as an expectation over $P(\phi_{(.)}(X),Y)$. 

The first term $\mathcal{H}(Y)$ is the entropy of the output random variable and is independent of the model. Therefore, we can ignore $\mathcal{H}(Y)$ if we only examine the difference in MI of two methods. Table \ref{tab:mi} shows the difference $\mathcal{I}(Y;\phi_{(.)}(X))-\mathcal{I}(Y;\phi_{SDT}(X))$ for our proposed CL methods. Here $\phi_{(.)}(X)$ denotes the concatenation of all hidden layer and final layer output representations of a domain-tuned (but not fine-tuned) model. To estimate the term $E_{p(\phi_{(.)}(X),Y)}[\log P(Y|\phi_{(.)}(X))]$, we train a two layer neural network on $(y_i,\phi_{(.)}(x_i))$ samples from MedNLI training data. The expectation is approximated by aggregating over $(y_i,\phi_{(.)}(x_i))$ samples from MedNLI test data. 

\begin{table*}
\begin{tabular}{p{0.2\textwidth}p{0.5\textwidth}p{0.2\textwidth}}
\hline
Dataset Source & Use & Number of words \\ \hline \hline
Pubmed +MIMIC-III & As \textit{Target} domain dataset for Pubmed+Mimic experiments & 2,454,355,020\\
MIMIC-III &  As \textit{Target} domain dataset for Clinical-only experiments & 516,586,695\\
WikiText & \textit{Source} domain dataset used for CL regularization in RH,EWC,GEM and DIS & 538,061,015 \\ \hline
\end{tabular}

\caption{The Dataset size for all datasets used for domain-tuning training and CL regularization are provided. }
\label{tab:domain_tuning_datasets}
\end{table*}

\section{Dataset Details}
\label{sec:appendix_b}
\subsection{Domain-Tuning Datasets}
We use a block randomized sub-section of WikiText corpus as a substitute for our \textit{source} domain text. The main domain-tuning experiments were done on Pubmed+MIMIC. We use a subset of the total Pubmed+MIMIC as our \textit{target} domain dataset, due to resource contraints. We also performed experiment on a smaller \textit{target} domain set comprised on MIMIC-III documents only. The size for each dataset is provided in Table \ref{tab:domain_tuning_datasets}.

Our choice of clinical+ medical dataset was based on :
\begin{itemize}
    \item Pubmed and MIMIC-III datasets are openly available
    \item Widely used tasks such as NLI and QA are available for clinical+ medical target domain.
\end{itemize}

Other possible domains such as scientific and law, either did not have open domain datasets or did not have \textit{target} domain QA and SNLI datasets.

\subsection{Downstream Datasets}
For the question answering datasets, SQuAD and emrQA, we used a doc stride of $128$ and a window size of $384$ across all the datapoints for each model\footnote{These are default values from \url{https://github.com/huggingface/transformers} implementation}. This results in upsampling of certain question answer pairs with different context passage windows. We also reject question answer pairs where the answer is not within the context size. For SNLI, the datapoints with `-' as their gold label were ignored resulting in slightly fewer datapoints after processing. These processing steps results in different pre and post processed dataset size. These statistics are presented in Table~\ref{tab:datapoints}. 

We use the standard train/test set defined for SNLI and MedNLI . For SQuAD 2.0 we use the released dev set as the test set. For EMRQA, we split the documents into train and test. We then extract Question, Answer and Context from relevant documents to obtain the train/test QA set. The train split contains 602 documents and the test set contains 151 documents.

\begin{figure*}[ht]
\centering
\includegraphics[width=0.95\textwidth]{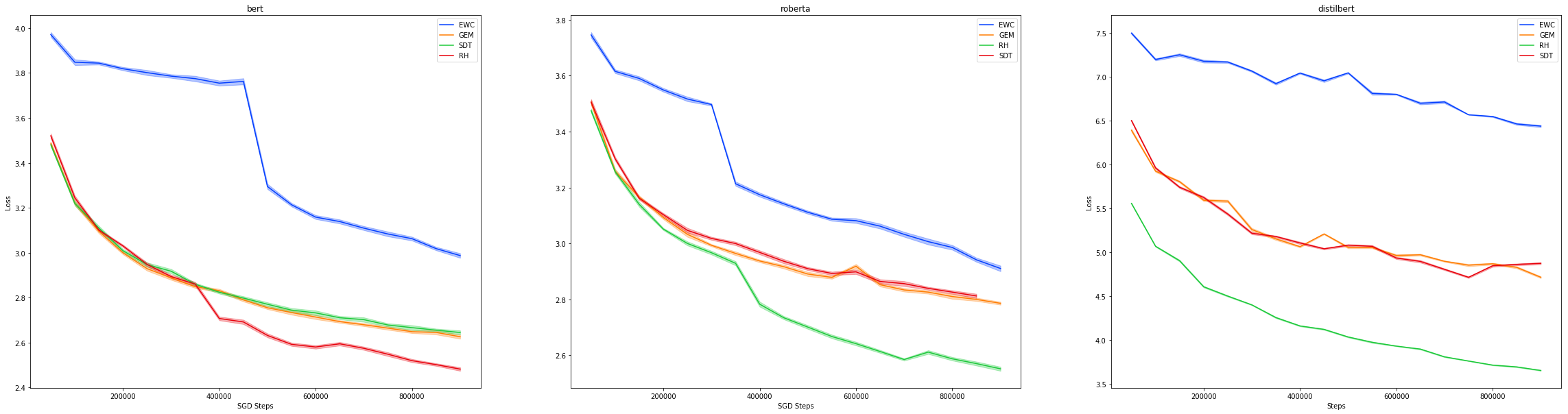}
\caption{Self-supervised loss $\mathcal{L_T}$ on \textit{target} domain validation set for all models. L2 model's loss is very high, and is not included in the graph to show higher y-axis resolution for other CL models.}
\end{figure*}

\begin{figure*}[h]
\centering
\includegraphics[width=0.95\textwidth]{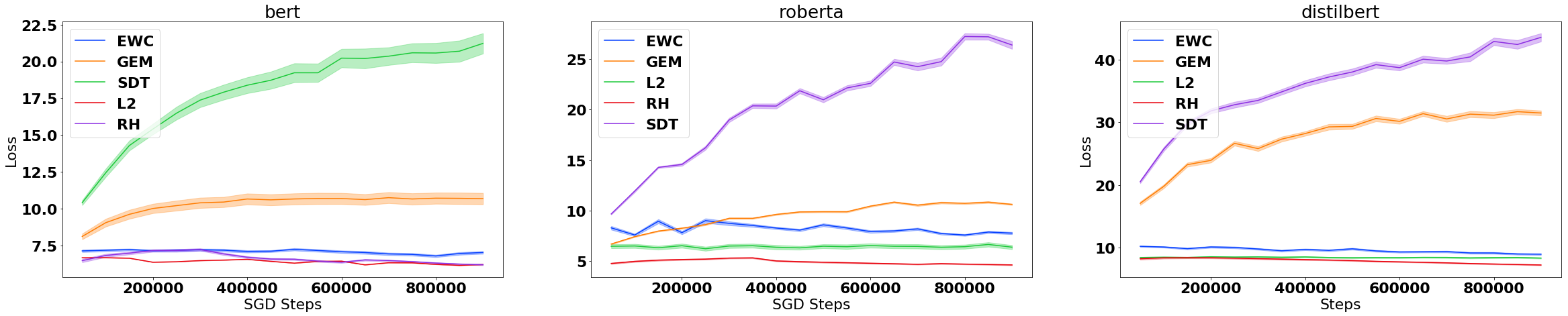}
\caption{Self-supervised loss $\mathcal{L_S}$ on \textit{source} domain validation set for all CL models.}
\end{figure*}

\begin{table}[!htbp]
\resizebox{\linewidth}{!}{%
\begin{tabular}{l|rr|rr}
\toprule
\textbf{}      & \multicolumn{2}{c|}{\textbf{Pre-processed}}                                         & \multicolumn{2}{c}{\textbf{Post-processed}}                                            \\ \toprule
\textbf{Dataset} & \multicolumn{1}{c}{\textbf{\# Train}} & \multicolumn{1}{c|}{\textbf{\# Test}} & \multicolumn{1}{c}{\textbf{\# Train}} & \multicolumn{1}{c}{\textbf{\# Test}} \\ \midrule
\textbf{SQuAD}   & 130,319                                 & 11,873                                 & 135,228                                 & 12,661                                 \\ 
\textbf{emrQA}   & 262,998                                 & 61,398                                 & 280,888                                 & 66,457                                 \\ 
\textbf{SNLI}    & 560,151                                 & 9,999                                 & 559,208                                 & 9,823                                  \\ 
\textbf{MedNLI}  & 12,626                                  & 1,421                                  & 12,626                                  & 1,421                                  \\ \bottomrule
\end{tabular}
}
\caption{Number of original and post-processed datapoints for each dataset.}
\label{tab:datapoints}
\end{table}

\section{Additional Experiments}
\label{sec:add_exp}
An expanded version of Table \ref{tab:results} with Distillation CL method and standard deviations are provided in Table \ref{tab:full_results}.

We also performed downstream task experiments on a smaller \textit{target} domain using MIMIC-III documents as the \textit{target} domain text source. The results are available  in Table \ref{tab:mimic_results}.

\begin{table*}
\centering
\resizebox{\linewidth}{!}{%
\begin{tabular}{cc|ccc|ccc}
\hline
& & \multicolumn{3}{c|}{\textbf{Downstream tasks related to \textit{Source} domain}} & \multicolumn{3}{c|}{\textbf{Downstream tasks related to \textit{Target} domain}}\\
\hline
& & \multicolumn{2}{c}{\bf SQuAD 2.0} & \bf SNLI & \multicolumn{2}{c}{\bf EMRQA} & \bf MedNLI\\
& & EM & F-score & Accuracy & EM & F-score & Accuracy\\
\hline
\multirow{6}{*}{\rotatebox{90}{\textsc{bert}}}
& \bf NDT & 71.78 $\pm$ 0.36 & 75.5 $\pm$ 0.32 & 90.45 $\pm$ 0.13 & 74.86 $\pm$ 0.28 & 80.43 $\pm$ 0.3 & 78.49 $\pm$ 0.56 \\
& \bf SDT & 71.25 $\pm$ 0.25 & 74.63 $\pm$ 0.3 & 88.45 $\pm$ 0.16 & 76.3 $\pm$ 0.21 & 81.86 $\pm$ 0.12 & 79.92 $\pm$ 0.46 \\
& \bf RH & 70.76 $\pm$ 0.26 & 74.41 $\pm$ 0.29 & 89.17 $\pm$ 0.05 & 76.26 $\pm$ 0.23 & 81.86 $\pm$ 0.28 & 80.46 $\pm$ 0.57 \\
& \bf EWC & 71.11 $\pm$ 0.33 & 74.81 $\pm$ 0.31 & 90.14 $\pm$ 0.03 & 75.8 $\pm$ 0.26 & 81.29 $\pm$ 0.18 & 80.93 $\pm$ 0.3 \\
& \bf GEM &  68.88 $\pm$ 0.34 & 72.43 $\pm$ 0.31 & 88.37 $\pm$ 0.2 & 75.9 $\pm$ 0.19 & 81.71 $\pm$ 0.14 & 78.47 $\pm$ 0.45 \\
& \bf L2 & 72.18 $\pm$ 0.49 & 75.86 $\pm$ 0.52 & 90.4 $\pm$ 0.12 & 75.76 $\pm$ 0.3 & 80.96 $\pm$ 0.13 & 78.14 $\pm$ 1.03 \\
& \bf DIS & 69.6 $\pm$ 0.57 & 73.27 $\pm$ 0.51 & 88.72 $\pm$ 0.08 & 75.95 $\pm$ 0.04 & 81.47 $\pm$ 0.08 & 80.51 $\pm$ 0.53 \\
\hline
\multirow{6}{*}{\rotatebox{90}{Ro\textsc{bert}a}} 
& \bf NDT & 78.16 $\pm$ 0.23 & 81.79 $\pm$ 0.15 & 91.38 $\pm$ 0.03 & 74.57 $\pm$ 0.07 & 80.45 $\pm$ 0.18 & 81.04 $\pm$ 0.34 \\
& \bf SDT & 72.61 $\pm$ 0.17 & 76.34 $\pm$ 0.17 & 89.59 $\pm$ 0.12 & 74.24 $\pm$ 0.25 & 80.09 $\pm$ 0.18 & 83.98 $\pm$ 0.39 \\
& \bf RH & 75.61 $\pm$ 0.15 & 79.31 $\pm$ 0.16 & 90.75 $\pm$ 0.17 & 74.93 $\pm$ 0.26 & 80.7 $\pm$ 0.11 & 85.43 $\pm$ 0.01 \\
& \bf EWC & 77.67 $\pm$ 0.16 & 81.44 $\pm$ 0.19 & 91.3 $\pm$ 0.07 & 74.98 $\pm$ 0.26 & 80.81 $\pm$ 0.3 & 83.65 $\pm$ 0.24 \\
& \bf GEM & 72.75 $\pm$ 0.09 & 76.57 $\pm$ 0.13 & 89.87 $\pm$ 0.24 & 74.27 $\pm$ 0.54 & 80.11 $\pm$ 0.27 & 84.43 $\pm$ 0.2 \\
& \bf L2 & 78.25 $\pm$ 0.09 & 81.84 $\pm$ 0.06 & 91.44 $\pm$ 0.01 & 74.68 $\pm$ 0.22 & 80.47 $\pm$ 0.21 & 80.44 $\pm$ 0.4 \\
& \bf DIS & 74.7 $\pm$ 0.29 & 78.41 $\pm$ 0.12 & 90.59 $\pm$ 0.08 & 74.4 $\pm$ 0.23 & 80.45 $\pm$ 0.08 & 83.91 $\pm$ 0.24 \\
\hline
\multirow{6}{*}{\rotatebox{90}{Distil\textsc{bert}}}
& \bf NDT & 65.47 $\pm$ 0.16 & 69.18 $\pm$ 0.17 & 89.69 $\pm$ 0.07 & 73.56 $\pm$ 0.12 & 79.3 $\pm$ 0.28 & 76.07 $\pm$ 0.19 \\
& \bf SDT & 64.88 $\pm$ 0.06 & 68.4 $\pm$ 0.14 & 87.25 $\pm$ 0.29 & 74.21 $\pm$ 0.07 & 80.26 $\pm$ 0.04 & 78.27 $\pm$ 0.18 \\
& \bf RH & 65.14 $\pm$ 0.33 & 68.89 $\pm$ 0.35 & 88.87 $\pm$ 0.1 & 74.35 $\pm$ 0.04 & 80.2 $\pm$ 0.05 & 78.46 $\pm$ 0.94 \\
& \bf EWC & 65.22 $\pm$ 0.34 & 69.01 $\pm$ 0.29 & 89.47 $\pm$ 0.07 & 73.71 $\pm$ 0.4 & 79.5 $\pm$ 0.28 & 76.26 $\pm$ 0.24 \\
& \bf GEM &  64.17 $\pm$ 0.11 & 67.78 $\pm$ 0.15 & 87.98 $\pm$ 0.06 & 74.26 $\pm$ 0.18 & 80.05 $\pm$ 0.02 & 78.58 $\pm$ 0.29 \\
& \bf L2 &  65.27 $\pm$ 0.55 & 68.92 $\pm$ 0.48 & 89.67 $\pm$ 0.22 & 73.57 $\pm$ 0.09 & 79.38 $\pm$ 0.09 & 75.81 $\pm$ 0.32 \\
& \bf DIS & 64.98 $\pm$ 0.13 & 68.72 $\pm$ 0.05 & 88.71 $\pm$ 0.12 & 74.2 $\pm$ 0.26 & 80.03 $\pm$ 0.2 & 78.11 $\pm$ 0.44 \\
\hline
\end{tabular}
}
\caption{\label{tab:full_results}
Downstream task results with standard deviations from our experiments across all models and techniques for \textit{target} domain. CL domain-tuned models are finetuned for each task to obtain the results. All results are averaged over three runs. NDT refers to the base model that is not domain tuned.}
\end{table*}

\begin{landscape}

\begin{table}
\centering
\begin{tabular}{cc|ccc|ccc}
\hline
& & \multicolumn{3}{c|}{\textbf{Downstream tasks related to \textit{Source} domain}} & \multicolumn{3}{c|}{\textbf{Downstream tasks related to \textit{Target} domain}}\\
\hline
& & \multicolumn{2}{c}{\bf SQuAD 2.0} & \bf SNLI & \multicolumn{2}{c}{\bf EMRQA} & \bf MedNLI\\
& & EM & F-score & Accuracy & EM & F-score & Accuracy\\
\hline
\multirow{7}{*}{\rotatebox{90}{\textsc{bert}}}
& \bf NDT & 71.83 $\pm$ 0.12 & 75.59 $\pm$ 0.04 & 89.89 $\pm$ 0.06 & 74.84 $\pm$ 0.03 & 80.51 $\pm$ 0.18 & 78.65 $\pm$ 0.5 \\ 
& \bf SDT & 65.92 $\pm$ 0.12 & 69.64 $\pm$ 0.06 & 88.38 $\pm$ 0.24 & 76.04 $\pm$ 0.22 & 81.62 $\pm$ 0.18 & 80.15 $\pm$ 0.5 \\ 
& \bf RH & 69.16 $\pm$ 0.15 (3.24)  & 72.77 $\pm$ 0.13 (3.13)  & 88.72 $\pm$ 0.39 (0.34)  & 75.63 $\pm$ 0.28 (-0.41)  & 81.32 $\pm$ 0.13 (-0.30)  & 81.21 $\pm$ 0.47 (1.06)  \\ 
& \bf EWC & 72.16 $\pm$ 0.12 (6.24)  & 75.79 $\pm$ 0.16 (6.15)  & 89.88 $\pm$ 0.12 (1.50)  & 75.08 $\pm$ 0.19 (-0.96)  & 80.43 $\pm$ 0.12 (-1.19)  & 80.58 $\pm$ 0.62 (0.43)  \\ 
& \bf GEM & 67.31 $\pm$ 0.12 (1.39)  & 71.1 $\pm$ 0.02 (1.46)  & 88.8 $\pm$ 0.08 (0.42)  & 75.83 $\pm$ 0.3 (-0.21)  & 81.5 $\pm$ 0.2 (-0.12)  & 80.63 $\pm$ 0.31 (0.48)  \\ 
& \bf L2 & 71.79 $\pm$ 0.53 (5.87)  & 75.51 $\pm$ 0.53 (5.87)  & 89.88 $\pm$ 0.11 (1.50)  & 75.0 $\pm$ 0.07 (-1.04)  & 80.5 $\pm$ 0.1 (-1.12)  & 78.87 $\pm$ 0.18 (-1.28)  \\ 
& \bf DIS & 66.83 $\pm$ 0.3 (0.91)  & 70.54 $\pm$ 0.34 (0.90)  & 88.66 $\pm$ 0.21 (0.28)  & 75.89 $\pm$ 0.25 (-0.15)  & 81.42 $\pm$ 0.29 (-0.20)  & 81.02 $\pm$ 0.09 (0.87)  \\
\hline
\multirow{7}{*}{\rotatebox{90}{Ro\textsc{bert}a}} 
& \bf NDT & 78.18 $\pm$ 0.14 & 81.76 $\pm$ 0.1 & 90.8 $\pm$ 0.07 & 74.76 $\pm$ 0.17 & 80.53 $\pm$ 0.19 & 80.98 $\pm$ 0.44 \\ 
& \bf SDT & 75.11 $\pm$ 0.04 & 78.77 $\pm$ 0.02 & 90.37 $\pm$ 0.2 & 75.2 $\pm$ 0.14 & 80.86 $\pm$ 0.21 & 84.47 $\pm$ 0.38 \\ 
& \bf RH & 75.9 $\pm$ 0.25 (0.79)  & 79.65 $\pm$ 0.23 (0.88)  & 90.6 $\pm$ 0.07 (0.23)  & 74.71 $\pm$ 0.1 (-0.49)  & 80.33 $\pm$ 0.17 (-0.53)  & 84.35 $\pm$ 0.35 (-0.12)  \\ 
& \bf EWC & 78.1 $\pm$ 0.03 (2.99)  & 81.73 $\pm$ 0.09 (2.96)  & 90.76 $\pm$ 0.01 (0.39)  & 74.95 $\pm$ 0.27 (-0.25)  & 80.85 $\pm$ 0.22 (-0.01)  & 83.39 $\pm$ 0.3 (-1.08)  \\ 
& \bf GEM & 75.21 $\pm$ 0.04 (0.10)  & 78.86 $\pm$ 0.04 (0.09)  & 90.34 $\pm$ 0.11 (-0.03)  & 74.9 $\pm$ 0.36 (-0.30)  & 80.65 $\pm$ 0.18 (-0.21)  & 84.82 $\pm$ 0.14 (0.35)  \\ 
& \bf L2 & 78.22 $\pm$ 0.04 (3.11)  & 81.65 $\pm$ 0.08 (2.88)  & 90.56 $\pm$ 0.09 (0.19)  & 74.41 $\pm$ 0.17 (-0.79)  & 80.29 $\pm$ 0.1 (-0.57)  & 79.33 $\pm$ 0.49 (-5.14)  \\ 
& \bf DIS & 75.41 $\pm$ 0.13 (0.30)  & 79.0 $\pm$ 0.1 (0.23)  & 90.95 $\pm$ 0.05 (0.58)  & 75.38 $\pm$ 0.12 (0.18)  & 80.82 $\pm$ 0.15 (-0.04)  & 84.64 $\pm$ 0.26 (0.17)  \\
\hline
\multirow{7}{*}{\rotatebox{90}{Distil\textsc{bert}}}
& \bf NDT & 65.14 $\pm$ 0.3 & 68.78 $\pm$ 0.41 & 89.51 $\pm$ 0.13 & 73.02 $\pm$ 0.08 & 78.76 $\pm$ 0.1 & 76.26 $\pm$ 0.29 \\ 
& \bf SDT & 61.72 $\pm$ 0.22 & 65.57 $\pm$ 0.19 & 88.43 $\pm$ 0.1 & 74.37 $\pm$ 0.07 & 80.17 $\pm$ 0.04 & 78.47 $\pm$ 0.32 \\ 
& \bf RH & 63.83 $\pm$ 0.13 (2.11)  & 67.45 $\pm$ 0.07 (1.88)  & 88.97 $\pm$ 0.26 (0.54)  & 74.77 $\pm$ 0.24 (0.40)  & 80.47 $\pm$ 0.23 (0.30)  & 78.98 $\pm$ 0.16 (0.51)  \\ 
& \bf EWC & 64.86 $\pm$ 0.25 (3.14)  & 68.53 $\pm$ 0.28 (2.96)  & 89.07 $\pm$ 0.05 (0.64)  & 74.37 $\pm$ 0.21 (0.00)  & 80.19 $\pm$ 0.19 (0.02)  & 78.28 $\pm$ 0.07 (-0.19)  \\ 
& \bf GEM & 62.37 $\pm$ 0.23 (0.65)  & 66.21 $\pm$ 0.1 (0.64)  & 88.75 $\pm$ 0.07 (0.32)  & 74.62 $\pm$ 0.22 (0.25)  & 80.32 $\pm$ 0.24 (0.15)  & 78.56 $\pm$ 0.35 (0.09)  \\ 
& \bf L2 & 65.12 $\pm$ 0.43 (3.40)  & 68.79 $\pm$ 0.44 (3.22)  & 89.52 $\pm$ 0.07 (1.09)  & 73.7 $\pm$ 0.09 (-0.67)  & 79.51 $\pm$ 0.16 (-0.66)  & 76.33 $\pm$ 0.15 (-2.14)  \\ 
& \bf DIS & 62.36 $\pm$ 0.3 (0.64)  & 66.16 $\pm$ 0.19 (0.59)  & 88.71 $\pm$ 0.14 (0.28)  & 74.31 $\pm$ 0.18 (-0.06)  & 79.96 $\pm$ 0.13 (-0.21)  & 78.94 $\pm$ 0.26 (0.47)  \\ 
\hline
\end{tabular}
\caption{\label{tab:mimic_results}
Downstream task results from our experiments across all models and techniques for ``mimic-only'' \textit{target} domain.}
\end{table}

\end{landscape}

\end{document}